\documentclass[11pt]{article}
\usepackage{acl2012}
\usepackage{times}
\usepackage{latexsym}
\usepackage{amsmath}
\usepackage{amssymb}			
\usepackage{multirow}
\usepackage{url}
\usepackage{booktabs}			
\usepackage{paralist}
\usepackage{amsbsy}   
\usepackage{color}
\usepackage{footnote}
\usepackage{bold-extra}
\usepackage{comment}
\usepackage{tikz}
\usetikzlibrary {positioning}
\usetikzlibrary{arrows}

\title{Studying the Wikipedia Hyperlink Graph\\for Relatedness and Disambiguation\\
}

\author{Eneko Agirre \\
  IXA NLP group \\
  UPV/EHU \\
  {\tt e.agirre@ehu.eus} \\\And
  Ander Barrena \\
  IXA NLP group \\
  UPV/EHU \\
  {\tt ander.barrena@ehu.eus} \\\And
  Aitor Soroa \\
  IXA NLP group \\
  UPV/EHU \\
  {\tt a.soroa@ehu.eus}
}

\date{}

\begin{document}
\maketitle
\begin{abstract}
  Hyperlinks and other relations in Wikipedia are a extraordinary resource
  which is still not fully understood. In this paper we study the
  different types of links in Wikipedia, and contrast the use of the
  full graph with respect to just direct links. We apply a well-known
  random walk algorithm on two tasks, word relatedness and
  named-entity disambiguation.  We show that using the full graph is
  more effective than just direct links by a large margin, that
  non-reciprocal links harm performance, and that there is no benefit
  from categories and infoboxes, with coherent results on both
  tasks. We set new state-of-the-art figures for systems based on
  Wikipedia links, comparable to systems exploiting several
  information sources and/or supervised machine learning. Our approach
  is open source, with instruction to reproduce results, and amenable to be
  integrated with complementary text-based methods.

\end{abstract}

\section{Introduction}
\label{sec:introduction}

\begin{figure}[t]
  \hspace{-25pt}\includegraphics[scale=0.70]{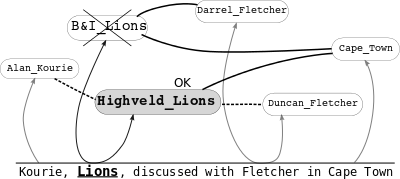}
    \caption{Simplified example motivating the use of the full
      graph. It shows the disambiguation of \emph{Lions} in
      ``\emph{Alan Kourie, CEO of the Lions franchise, had discussions
        with Fletcher in Cape Town}''. Each mention is linked to the
      candidate entities by arrows, e.g. \texttt{B\&I\_Lions} and
      \texttt{Highveld\_Lions}  for
      \emph{Lions}. Solid lines correspond to direct hyperlinks and
      dashed lines to a path of several links. An algorithm using
      direct links alone would incorrectly output
      \texttt{B\&I\_Lions}, while one using the full graph would
      correctly choose \texttt{Highveld\_Lions}. }
  \label{fig:example-show-disamb}
\end{figure}

Hyperlinks and other relations between concepts and instances in Wikipedia
 have been successfully used in semantic tasks
\cite{milne_open-source_2013}. Still, many questions about the best
way to leverage those links remain unanswered. For instance, methods
using direct hyperlinks alone would wrongly disambiguate \emph{Lions} in
Figure \ref{fig:example-show-disamb} to \texttt{B\&I\_Lions}, a rugby
team from Britain and Ireland, as it shares two direct links to
potential referents  in the context  (\emph{Darrel
  Fletcher}, a British football player, and \emph{Cape Town}, the city
where the team suffered some memorable defeats), while
\texttt{Highveld\_Lions}, a cricket team from South Africa, has only 
one. When considering the whole graph of hyperlinks we find that the
cricket team is related to two cricketers named
\emph{Alan Kourie} and \emph{Duncan Fletcher} and could thus pick the right
entity for \emph{Lions} in this context. In this paper we will study
this and other questions about the use of hyperlinks in word
relatedness \cite{gabrilovich2007csr} 
and named-entity disambiguation,
NED \cite{Hachey2012}.

Previous work on this area has typically focused on novel algorithms which work on a
specific mix of resource, information source, task and test dataset
(cf. Sect. \ref{sec:comp-relat-work}). In the case of NED, the
evaluation of the disambiguation component is  confounded by
interactions with mention spotting and candidate generation. With very
few exceptions, there is little analysis of components and
alternatives, and it is very difficult to learn any insight beyond the
fact that the mix under study attained certain performance on the target
dataset\footnote{See \cite{Hachey2012} and \cite{DBLP:journals/jair/GarciaAF14} for two exceptions on NED. The first is limited to a single dataset, the second explores methods based on direct links, which we extend to using the full graph.}. The number
of algorithms and datasets is growing by the day, with no
well-established single benchmark, and the fact that some systems are
developed on test data, coupled with reproducibility problems \cite[on
word relatedness]{fokkens-EtAl:2013:ACL2013}, makes it very difficult
to know where the area stands. There is a need for clear points
of reference which allow to understand where each information source and
algorithm stands with respect to other alternatives.

We thus depart from previous work, seeking to set such a point of reference, and
focus on a single knowledge source (hyperlinks in Wikipedia) with a clear
research objective: given a well-established random walk algorithm
(Personalized PageRank~\cite{have02}) we explore sources of links and
filtering methods, and contrast the use of the full graph with respect
to using just direct links. We follow a clear
development/test/analysis methodology, evaluating on a extensive range
of both relatedness and NED datasets. The results are confirmed in
both tasks, yielding more support to the findings in this
research. All software and data are publicly available, with instructions to obtain 
out-of-the-box replicability\footnote{\url{http://ixa2.si.ehu.es/ukb/README.wiki.txt}}.

The contributions of our research are the following: (1) We show for the
first time that performing random walks over the full graph is preferable
than considering only direct links. (2) We study several sources of links,
showing that non-reciprocal links hurt and that the contribution of the
category structure and links in infoboxes is residual. (3) We set the new
state-of-the-art for systems based on Wikipedia links for \emph{both}
word relatedness and named-entity disambiguation.  The results are close to
the best systems to date, which use several information sources and/or
supervised machine learning techniques, and specialize on either relatedness
or disambiguation. Our work shows that a careful analysis of varieties of
graphs using a well-known random walk algorithm pays off more than most
ad-hoc algorithms.

The article is structured as follows. We first present previous work,
followed by the different options to build hyperlink graphs. Sect.
\ref{sec:pager-pers-pager} reviews random walks for relatedness and
NED. Sect. \ref{sec:note-exper-meth} sets the experimental
methodology, followed by the analysis and results on development data
(Sect. \ref{sec:studying-graph}) and the comparison to the
state of the art (Sect. \ref{sec:comp-relat-work}). Finally, Sect.
\ref{sec:concl-future-work} draws the conclusions.

\section{Previous work}
\label{sec:previous-work}

The irruption of Wikipedia has opened up enormous opportunities for
natural language processing \cite{Hovy:2013:CBS:2405838.2405907}, with
many derived knowledge-bases, including
DBpedia 
\cite{Bizer:2009:DCP:1640541.1640848},
Freebase 
\cite{Bollacker:2008:FCC:1376616.1376746}, 
and BabelNet \cite{NavigliPonzetto:12aij}, to name a few. 
These resources have been successfully used on semantic processing tasks
like word relatedness, named-entity disambiguation (NED), also known as
entity linking, and the closely related Wikification.  Broadly speaking,
Wikipedia-based approaches to those tasks can be split between those using
the text in the articles (e.g., Gabrilovich and Markovitch,
2007) 
and those using the links between articles (e.g., Guo et al.,
2011). 

Relatedness systems take  
two words and return a high number
if the two words are similar or closely related\footnote{Relatedness
  is more general than similarity. For the sake of simplicity, we will
  talk about relatedness on this paper.} (e.g. \textsl{professor} - \textsl{student}),
and a low number otherwise (e.g. \textsl{professor} -
\textsl{cucumber}). Evaluation is performed comparing the returned
values to those by humans \cite{rubenstein1965ccs}.

In NED \cite{Hachey2012} the input is 
a mention of a named-entity in context 
and the output is the appropriate instance from Wikipedia, 
DBpedia or Freebase (cf. Figure \ref{fig:example-show-disamb}). 
Wikification is similar
\cite{mihalcea2007wikify}, but target terms include common nouns and
only relevant terms are disambiguated. Note that the disambiguation
component in Wikification and NED can be the same. 

Our work focuses on relatedness and NED. We favored NED over 
Wikification because of the larger number of systems and
evaluation datasets, but our conclusions are  applicable to
Wikification, as well as other Wikipedia-derived resources.

In this section we will focus on previous work using Wikipedia links
for relatedness, NED and Wikification. Although relatedness and
disambiguation are closely related (relatedness to context terms is an
important disambiguation clue for NED), most of the systems are evaluated 
in either relatedness or NED, with few exceptions, like 
WikiMiner \cite{milne_open-source_2013}, KORE
\cite{hoffart_kore:_2012} and the one presented in this paper.

Milne and Witten \shortcite{Milne08aneffective} are the first to use
hyperlinks between articles for relatedness. They compare two articles
according to the number of incoming links that they have in common
(i.e. overlap of direct-links) based on Normalized Google Distance (NGD), combined with several heuristics and collocation strength. 
In later work \cite{milne_open-source_2013}, they incorporated machine
learning.  The authors also apply their technique to NED
\cite{Milne2008}, using their relatedness measures to train a
supervised classifier.  Unfortunately they do not present results of
their link-based method alone, so we decided to reimplement it
(cf. Sect. \ref{sec:studying-graph}). We show that, under the same
conditions, using the full-graph is more effective in both tasks. We
also run their out-of-the-box
system\footnote{\url{https://sourceforge.net/projects/wikipedia-miner/}}
on the same datasets as ours (cf. Sect. \ref{sec:comp-relat-work}), with results below ours.

Apart from hyperlinks between articles, other works on relatedness use
the category structure
\cite{strube2006wikirelate,ponzetto07a,ponzetto11a} to run path-based
relatedness algorithms which had been successful on WordNet
\cite{Pedersen:2004:WMR:1614025.1614037}, or use relations in
infoboxes \cite{DBLP:journals/ai/NastaseS13}. In all cases, they
obtain performance figures well below hyperlink-based systems
(cf. Sect. \ref{sec:comp-relat-work}). We will explore the
contribution of such relations  (cf. Sect.
\ref{subsec:build-wikip-graphs}), incorporating them to the hyperlink
graph.

Attempts to use the whole graph of hyperlinks for relatedness have
been reported before.  Yeh et
al. \shortcite{yeh-EtAl:2009:TextGraphs4} obtained very low results on
relatedness using an algorithm based on random walks similar to
ours. Similar in spirit, Yazdani and Popescu-Belis \shortcite{Yazdani:2013:CTS:2405838.2405916} built
a graph derived from the Freebase Wikipedia Extraction dataset, which
is derived but richer than Wikipedia.
Even if they mix hyperlinks with textual similarity, their results are
lower than ours. One of the key differences with these systems is that
we remove non-reciprocal links (cf. Sect.
\ref{subsec:build-wikip-graphs}).

Regarding link-based methods for NED, there is only one system which
relies exclusively on hyperlinks. Guo et al. \shortcite{guo_graph-based_2011} use
direct hyperlinks between the target entity and the mentions in the
context, counting the number of such links. We show that the use of the full graph produces better results.  

The rest of NED systems present complex
combinations.  Lemahnn et al. \shortcite{Lehmann2010} present a
supervised system combining features based on hyperlinks, categories,
text similarity and relations from infoboxes. Despite their complex
and rich system, we will show that they perform worse than our system. \newcite{Hachey:2011:GNE:2050963.2050980} explored hyperlinks
beyond direct links for NED, building subgraphs for each context using
paths of length two departing from the context terms, combined with
text-based relatedness. We will show that the full graph is more
effective than limiting the distance to two, and report better results
than their system. Several authors have included direct links using the aforementioned NGD
in their combined systems \cite{RatinovRDA11,Hoffart11}. Unfortunately, they do no report separate results for
the NGD component.
In very recent work \newcite{DBLP:journals/jair/GarciaAF14} compare
NGD with several other algorithms using direct links, but do not
explore the full graph, or try to characterize links. We will see that
their results are well below ours (cf. Sect.
\ref{sec:comp-relat-work}).

Graph-based algorithms for relatedness and disambiguation have been
successfully used on other resources, particularly WordNet. Hughes and
Ramage \shortcite{hughes2007emnlp} were the first presenting a random walk
algorithm over the WordNet graph.  Agirre et al.  \shortcite{AGIRRE10.534}
improved over their results using a similar random walk algorithm on
several variations of WordNet relations, reporting the best results to
date among WordNet-based algorithms. The same algorithm
was used for word sense disambiguation
\cite{DBLP:journals/coling/AgirreLS14}, also reporting state-of-the-art results. We use the same open source
software in our experiments. As an alternative to random walks,
Tsatsaronis et al. \shortcite{DBLP:journals/jair/TsatsaronisVV10} use a
path-based system over the WordNet relation graph.  

In more recent work \cite{DBLP:conf/aaai/NavigliP12,Pilehvaretal:2013}, the authors present two relatedness algorithms 
 for BabelNet, an enriched version of WordNet including
articles from Wikipedia, hyperlinks and cross-lingual relations
from non-English Wikipedias.  In related work, Moro et
al. \shortcite{Moro:2014:ELmeetsWSD} present a multi-step NED algorithm on BabelNet,
building semantic graphs for each context.
We will show that Wikipedia hyperlinks alone are
able to provide similar performance on both tasks.

\section{Building Wikipedia  Graphs}
\label{subsec:build-wikip-graphs}

Wikipedia pages can be classified into main articles, category pages,
redirects and disambiguation pages. Given a Wikipedia dump (a snapshot from
April 4, 2013), we mine links between articles, between articles and
category pages, as well as the links between category pages (the category
structure). Our graphs include a directed edge from one article to another
iff the text of the first article contains a hyperlink to the second
article. In addition, we also include hyperlinks in infoboxes.

The graph contains two types of nodes (articles and categories) and
three types of directed edges: hyperlinks from article to article
(\textbf{\textsc{H}}), infobox links from article to article
(\textbf{\textsc{I}}), links from article to category and links from 
category to  category (\textbf{\textsc{C}}).

We constructed several graphs using different combinations of nodes and
edges. In addition to the directed versions (\textbf{\textsc{d}}) we also
constructed an undirected version (\textbf{\textsc{u}}), and a reduced graph
which only contains links which are reciprocal (\textbf{\textsc{r}}), that
is, we add a pair of edges between $a1$ and $a2$ if and only if there exists
a hyperlink from $a1$ to $a2$ and from $a2$ to $a1$. \textbf{Reciprocal}
links capture the intuition that both articles are relevant to each other,
and tackle issues with links to low relevance articles, e.g. links to
articles on specific years like 1984.
Some authors weight links according to their relevance
\cite{milne_open-source_2013}. Our heuristic to keep
only reciprocal links can be seen as a simpler, yet effective, method
to avoid low relevance links.  

\begin{table}[t]
  \small
  \centering
\begin{tabular}{rrr|l|l}
	Graph & Edges & Nodes & RG & TAC09$_{200}$\\
    \hline 
     \textsc{    Cd}        & 18,803K  & 4,873K & 51.1 $\dagger$ $\ddagger$& 49.5 $\dagger$  $\ddagger$  \\                   
     \textsc{    Cu}        & 37,598K  & 4,873K & 72.9 $\dagger$ $\ddagger$& 65.5 $\dagger$   $\ddagger$ \\ \hline            
     \textsc{    Id}        & 6,572K   & 1,860K & 43.1 $\dagger$ $\ddagger$& 57.0 $\dagger$   $\ddagger$ \\                   
     \textsc{    Iu}        & 12,692K  & 1,860K & 52.8 $\dagger$ $\ddagger$& 65.5 $\dagger$   $\ddagger$ \\ \hline            
     \textsc{    Hd}        & 90,674K  & 4,103K & 75.1 $\dagger$ $\ddagger$ & 65.0 $\dagger$   $\ddagger$ \\                   
     \textsc{   {Hu}}       & 165,258K & 4,103K &        {76.6} $\ddagger$ &        {66.0} $\dagger$ $\ddagger$  \\
     \textsc{\textbf{Hr}}   & 16,338K  & 2,955K & \textbf{88.4}            & \textbf{68.5}   \\ \hline            
     \textsc{       {HrCu}} & 53,005K  & 4,898K &        {78.2} $\ddagger$ &        {67.5} $\ddagger$ \\        
     \textsc{       {HrIu}} & 26,394K  & 3,273K &        {82.9} $\ddagger$ &        {68.0} $\ddagger$ \\ \hline 
     \textsc{HrCuIu}        & 63,184K  & 4,900K & 75.6 $\dagger$         $\ddagger$ & 67.5   $\ddagger$          \\                    
    \hline 
\end{tabular}
\caption{Statistics for selected  graphs and results on development data for relatedness (RG, Spearman) 
  and NED (TAC09$_{200}$, accuracy) with default parameters (see text). 
  See Sect. \ref{subsec:uniform-graph-based} for abbreviations. 
  $\dagger$ for stat. significant differences with \textsc{Hr} in either RG or TAC09$_{200}$. 
  $\ddagger$ for stat. signif. when comparing on all relatedness or NED datasets.   
}
\label{table:dev-graphs}
\end{table}

Table \ref{table:dev-graphs} gives the number of nodes and edges in
some selected graphs. The graph with less edges is
the one with reciprocal hyperlinks \textbf{\textsc{Hr}},
and the graphs with most edges are those with undirected edges, as
each edge is modeled as two directed edges\footnote{This was done in
  order to combine undirected and reciprocal edges, and could be
  avoided in other cases.}. The number of nodes is similar in all,
except for the infobox graphs (infoboxes are only available for a
few articles), and the reciprocal graph \textbf{\textsc{Hr}}, as
relatively few nodes have reciprocal edges.

\begin{table}[t]
\footnotesize
\centering
\begin{tabular}{lrr}
  Article         & Freq. & Prob.\\
  \hline
  \textsc{Gotham\_City} & 32 & 0.38\\
  \textsc{Gotham\_(magazine)} & 15 & 0.18 \\
  \ldots\\
  \textsc{New\_York\_City} & 1 & 0.01\\
  \textsc{Gotham\_Records} & 1 & 0.01\\
  \hline 
  \end{tabular}
  \caption{Partial view of dictionary entry for ``gotham''. The probability is calculated as the ratio between the frequency and the total count.
  }
  \label{table:dict}
\end{table}

\subsection{Building the dictionary}
\label{sec:building-dictionary}

In order to link running text to the articles in the graph, we use a
dictionary, i.e., a static association between string mentions with all
possible articles the mention can refer to.

We built our dictionary from the same Wikipedia dump, using article titles, redirections, disambiguation pages, and
anchor text. Mention strings are lowercased and all text between parentheses
is removed. If an anchor links to a disambiguation page, the text is
associated with all possible articles the disambiguation page points
to. Each association between a mention and article is scored with the
prior probability, estimated as the number of times that the mention
occurs in an anchor divided by the total number of
occurrences of the mention as anchor. Note that our dictionary can disambiguate
any mention, just returning the highest-scoring article. Table
\ref{table:dict} partially shows  a sample entry in our
dictionary. 

\begin{table}[t]
\scriptsize
\centering
\begin{tabular}{lrlr}
  \multicolumn{2}{c}{\footnotesize Drink} & \multicolumn{2}{c}{\footnotesize Alcohol}      \\
  \hline 
  \textsc{Drink }                 & .124 & \textsc{Alcohol }           & .145  \\
  \textsc{Alcoholic\_beverage}    & .036 & \textsc{Alcoholic\_beverage}& .026  \\
  \textsc{Drinking}               & .028 & \textsc{Ethanol }           & .018  \\
  \textsc{Coffee}                 & .020 & \textsc{Alkene}             & .006  \\
  \textsc{Tea }                   & .017 & \textsc{Alcoholism}         & .006  \\
  \hline 
  \end{tabular}
  \caption{Sample of the probability distribution returned by \textsc{ppr} for two words. Top five articles shown.}
  \label{table:ppv}
\end{table}

\section{Random Walks}
\label{sec:pager-pers-pager}

The PageRank random walk algorithm~\cite{brin98} is a method for ranking the
vertices in a graph according to their relative structural
importance. PageRank can be viewed as the result of a random walk process,
where the final rank of node $i$ represents the probability of a random walk
over the graph ending on node $i$, at a sufficiently large time.

Personalized PageRank (\textsc{ppr}) is a variation of
PageRank~\cite{have02}, where the query of the user defines the
importance of each node, biasing the resulting PageRank score to
prefer nodes in the vicinity of the query nodes.  The query bias is
also called the teleport vector. \textsc{ppr} has been successfully
used on the WordNet graph for relatedness
\cite{hughes2007emnlp,AGIRRE10.534} and WSD
\cite{Agirre:09a,DBLP:journals/coling/AgirreLS14}. In our experiments
we use UKB version 2.1\footnote{\url{http://ixa2.si.ehu.es/ukb}}, an
open source software for relatedness and disambiguation
based on \textsc{ppr}. For the sake of space, we will skip the
details, and refer the reader to those papers. \textsc{ppr} has two
parameters: the number of \textbf{iterations}, and the \textbf{damping
  factor}, which controls the relative weight of the teleport vector.

\subsection{Random walks on Wikipedia}
\label{subsec:uniform-graph-based}

Given a dictionary and graph derived from Wikipedia (cf. Sect.
\ref{subsec:build-wikip-graphs}), \textsc{ppr} expects a set of
mentions, i.e., a set of strings which can be linked to Wikipedia
articles via the dictionary. The method
first initializes the teleport vector: for each mention in the input, the
articles in the respective dictionary entry are set with an initial
probability, and the rest of articles are set to zero. We explored two options to set the initial
probability of each article: the uniform probability or
 the \textbf{prior} probability in the dictionary. When an article
appears in the dictionary entry for two mentions, the initial
probability is summed up. In a second step, we apply \textsc{ppr}
for a number of iterations, producing a probability distribution over
Wikipedia articles in the form of a \textsc{ppr} vector
(\textsc{ppv}).

The probability vector can be used for both  relatedness and
NED. For \textbf{relatedness} we produce a \textsc{ppv}
vector for each of the words to be compared, using the single word as
input mention. The relatedness between the target words is computed as
the cosine between the respective \textsc{ppv} vectors.
In order to speed up the computation, we can reduce the size of the
\textsc{ppv} vectors, setting to zero all values below rank $k$ after
ordering the values in decreasing order. 

Table \ref{table:ppv} shows the top 5 articles in the \textsc{ppv}
vectors of two sample words. The relatedness between pairs Drink and
Alcohol would be non-zero, as their respective vectors contain common
articles. 

For \textbf{NED} the input comprises the target entity mention and its
context, defined as the set of mentions occurring within a 101 token window
centered in the target. In order to extract mentions to articles in
Wikipedia from the context, we match the longest strings in our dictionary
as we scan tokens from left to right. We then initialize the teleport
probability with all articles referred by the mentions.  After computing
Personalized PageRank, we output the article with highest rank in
\textsc{ppv} among the possible articles for the target entity mention.
Figure \ref{fig:example-show-disamb} shows an example of NED.

If the prior is being used to initialize weights, we multiply the prior
probability with the Pagerank probabilities before computing the final
ranks. In the rare cases\footnote{Less than 3\% of instances.} where no
known mention is found in the context, we return the node with the highest
prior.

\begin{figure}[t]
\begin{compactenum}
\item \textbf{Graphs} in Table \ref{table:dev-graphs} (default: \textbf{Hr})
\item Number of \textbf{iterations} in PageRank\\ 
  $\boldsymbol i \in \{1,2,3,4,5,10,15 \ldots 50\}$ (default: \textbf{30})
\item \textbf{Damping factor} in PageRank:\\
  $\boldsymbol\alpha \in\{0.75, 0.8, 0.85, 0.90, 0.95, 0.99\}$ (default: \textbf{0.85})
\item Initializing with \textbf{prior} or not (\textbf{\textsc{P}} or
  \textbf{\textsc{$\boldsymbol\neg$P}}) (default: \textbf{P})
\item Relatedness: number of values in \textsc{ppv}: \\ $\boldsymbol k
  \in \{100, 200, 500, 1000, 2000, 5000, 10000\}$ (default: \textbf{5000})
\end{compactenum}
\caption{Summary of variants and parameters as well as the default values for each of them.} 
\label{fig:variants}
\end{figure}

Note that our NED and relatedness algorithms are related. NED is using
using relatedness, as Pagerank probabilities are capturing how related
is each candidate article to the context of the mention. Following the
first-order and second-order co-occurrence abstraction
\cite[Ch. 6]{islam2006second,Agirre:2007:WSD:1564561}, we can
interpret that we do NED using first-order relatedness,
while our relatedness uses second-order relatedness.

Figure \ref{fig:variants} summarizes all parameters mentioned so far, as
well as their default values, which were set following previous work
\cite{AGIRRE10.534,DBLP:journals/coling/AgirreLS14}.

\section{Experimental methodology}
\label{sec:note-exper-meth}

We summarize the datasets used in Table \ref{table:datasets}. RG, MC
and 353 are the most used relatedness datasets to date, with TSA and KORE being
more recent datasets where some top-ranking systems have been
evaluated. Word relatedness datasets were lemmatized and lowercased,
except for KORE, which is an entity relatedness dataset where the
input comprises article titles\footnote{We had to manually adjust the
  articles in KORE, as the exact title depends on the Wikipedia
  version. We missed 3 for our 2013 version, which could slightly
  degrade our results. 
}. Following common practice rank-correlation (Spearman) was used for
evaluation.

\begin{table}[t]
  \small
  \centering
  \begin{tabular}{llr}
  Name   & Reference                                & \#  \\        
  \hline
  RG     & \cite{rubenstein1965ccs}                 & 65 \\         
  MC     & \cite{doi:10.1080/01690969108406936}     & 30 \\         
  353    & \cite{gabrilovich2007csr}                & 353 \\        
  TSA    & \cite{Radinsky:2011:WTC:1963405.1963455} & 287 \\        
  KORE   & \cite{hoffart_kore:_2012}                & 420 \\        
  \hline
  TAC09  & \cite{McNamee2010}                       & 1675 \\       
  TAC10  & {\footnotesize \url{http://www.nist.gov/tac/}} & 1020 \\       
  TAC13  & {\footnotesize \url{http://www.nist.gov/tac/}} & 1183 \\       
  AIDA   & \cite{Hoffart11}                         & 4401 \\       
  KORE   & \cite{hoffart_kore:_2012}                & 143 \\        
  \hline
  \end{tabular}
  \caption{Summary of relatedness (top) and NED (bottom) datasets. Rightmost column for number of instances. 
     }
\label{table:datasets}
\end{table}

Regarding NED, the TAC Entity Linking competition is held
annually. Due to its popularity it is useful to set the
state of the art. We selected the datasets in 2009 and 2010, as they
have been used to evaluate several top ranking systems, as well as the
2013 dataset, which is the most recent. In addition, we also provide
results for AIDA, the largest and only dataset providing annotations
for all entities in the documents, and KORE, a recent, very small
dataset focusing on difficult mentions and short contexts. Evaluation
was performed using accuracy, the ratio between correctly
disambiguated instances and the total number of instances that have a
link to an entity in the knowledge base\footnote{Corresponds to
  {non-NIL accuracy} at TAC-KBP (also called {KB accuracy}) and {Micro
    P@1.0} in \cite{Hoffart11}}. Each dataset uses a different
Wikipedia version, but fortunately Wikipedia keeps redirects from
older article titles to the new version. As customary in the task, we
automatically map the articles returned by our system to the
version used in the gold standard.

Following standard practice in NED, we do not evaluate mention
detection\footnote{See \cite{cornolti_framework_2013} for a framework
  to evaluate both mention detection and disambiguation.}, that is,
the datasets already specify which are the target mentions. Note that
TAC provides so called ``queries'' which can be substrings of the full
mention, e.g. ``Smith'' for a mention like ``John Smith''). Given a
mention, we devised the following heuristics to improve candidate
generation: (1) remove substring contained in parenthesis from the
mention, then check dictionary, (2) if not found, remove ``the'' if
first token in the mention, then check dictionary, (3) if not found,
remove middle token if mention contains three tokens, then check
dictionary, (4) if not found, search for a matching entity using the
Wikipedia API\footnote{\url{http://en.wikipedia.org/w/api.php}}. The
heuristics provide an improvement of around 4 points on
development. Later analysis showed that these heuristics seem to be
only relevant on the TAC datasets, because of the way the query
strings are designed, but not on AIDA or KORE.

\subsection{Development and test}
\label{sec:development-test}

We wanted to follow a standard experimental design, with a clear
development/test split for each task. Unfortunately there is no
standard split in the literature, and the choice is difficult: The
development dataset should be representative enough to draw
conclusions on different alternatives and parameters, but at the same
time the most relevant datasets in the literature should be left for
testing, in order to have enough points for comparison.  In addition,
some recent algorithms suposedly setting the state of the art are only tested on
newly produced datasets. Note also that relatedness datasets are
small, making it difficult to find statistically significant
differences.

In order to strike a balance between the need for in-depth analysis
and fair comparison to previous results, we decided to focus on the
two oldest datasets from each task for development and analysis: RG
for relatedness and a subset of 200 polysemic instances from TAC09 for NED
(TAC09$_{200}$)\footnote{The dataset in \url{http://ixa2.si.ehu.es/ukb/README.wiki.txt} includes the subset.}. 
The rest will be used for test, where the parameters
have been set on development. Given the need for significant
conclusions, we re-checked the main conclusions drawn
from development data using the aggregation of all test datasets, but \textbf{only after}
the comparison to the state of the art had been performed. This way we ensure both a
fair comparison with the state of the art and a well-grounded analysis.

We performed significance tests using Fisher's z-transformation for
relatedness \cite[equation 14.5.10]{press2002numerical}, and paired
bootstrap resampling for NED~\cite{Nor89}, accepting differences with
p-value $<0.05$. Given the small size of the datasets, when necessary,
we also report statistical significance when joining all datasets as just mentioned.

\section{Studying the graph and parameters}
\label{sec:studying-graph}

In this section we study the performance of the different graphs and
parameters on the two development datasets, RG  and
TAC09$_{200}$. The next section reports the results on the
test sets for the best parameters, alongside state-of-the-art system results.

\begin{table}[t]
 \small
 \centering
  \begin{tabular}{l|cl|cl}
	Graph & Param. & RG & Param. & TAC09$_{200}$\\
    \hline 
        \textsc{Hr}  & \textbf{default} & \textbf{88.4} & \textbf{default}   & \textbf{68.5} \\            
        \textsc{Hr}  & $\neg$\textsc{P} & 87.0          & $\neg$\textsc{P}   & 49.0 $\dagger$\\   \hline   
        \textsc{Hr}  & $\alpha 0.85$    & 88.4          & $\alpha 0.85$      & 68.5 \\                     
        \textsc{Hr}  & $i 30$           & 88.4          & $i 15$             & 68.5 \\                     
        \textsc{Hr}  & $k 5000$         & 88.4          & --                 & -- \\                       
    \hline 
  \end{tabular}
  \caption{Parameters: Summary of results on development data for relatedness (RG, Spearman correlation) and NED (TAC09$_{200}$, accuracy) for several parameters using \textsc{Hr} graph. Parameters are set to default values (see text) except for the one noted explicitly. 
    $\dagger$ for statistical significant differences with respect to default.
  }
  \label{table:development}
\end{table}

As mentioned in Sect. \ref{subsec:uniform-graph-based}, \textsc{ppr}
has several parameters and variants (cf. Figure \ref{fig:variants}). We first checked exhaustively all
possible combinations for different graphs, with the rest of
parameters set to \textbf{default values}.
We then optimized each of the parameters in turn, seeking to answer the following questions:

\textbf{Which links help most?} Table \ref{table:dev-graphs} shows the
results for selected graphs. The first seven rows present the results
for each edge source in isolation, both using directed and undirected
edges. Categories and infoboxes suffer from producing smaller graphs,
with the hyperlinks yielding the best results. The undirected versions
improve over directed links in all cases, with the use of reciprocal
edges for hyperlinks obtaining the best results overall (the graphs
with reciprocal edges for categories and infoboxes were too small and
we omit them). The trend is the same in both relatedness and NED, 
highlighting the robustness of these results. 

Regarding combined graphs, we report the most significant
combinations. The reciprocal graph of hyperlinks outperforms all combinations (including
the combinations which were omitted), showing that categories and
infoboxes do not help or even degrade slightly the results.  The differences are
statistically significant (either on the individual datasets or in the aggregation on all datasets) in all cases,
confirming that \textsc{Hr} is significantly better.

The degradation or lack of improvement when using infoboxes is
surprising. We hypothesized that it could be caused by non-reciprocal
links in \textsc{HrIu}. In fact, removing non-reciprocal links from
\textsc{HrIu} improved results slightly on NED, matching those of
\textsc{Hr}. This lack of improvement with infoboxes, even when
removing non-reciprocal links, can be explained by the fact that only
5\% of reciprocal links in \textsc{Iu} are not in \textsc{Hr}. It
seems that this additional 5\% is not helping in this particular
dataset.  Regarding categories, the category structure is mostly a
tree, which is a structure where random walks do not seem to be
effective, as already observed in
\cite{DBLP:journals/coling/AgirreLS14} for WordNet.

\textbf{Is initialization of random walks important?} The second row
in Table \ref{table:development} reports the result when using uniform
distributions when initializing the random walks (instead of prior
probabilities). The results degrade in both datasets, the difference
being significant only for NED. This was later confirmed in the rest
of relatedness and NED datasets: using prior probabilities for
initialization improves results in all cases, but it is only
significant in NED datasets. These results show that relatedness is
less sensitive to changes in the distribution of meanings, that is,
using the more informative prior distributions of meaning only
improves results slightly. NED, on the contrary, is more sensitive, as
the distribution of senses affects dramatically the performance.

\textbf{Is the value of $\alpha$ and $i$ important? } The best
$\alpha$ on both datasets was obtained with  default values (cf. Table \ref{table:development}), in
agreement with related work
using WordNet 
\cite{AGIRRE10.534}. 
The lowest number of
iterations where convergence was obtained were 30 and 15,
respectively, although as few as 5 iterations yielded very similar
performance (87.1 on relatedness, 68.0 on NED). 

\textbf{Is the size of the vector, $k$, important for relatedness?} The
best performance was attained for the default $k$, with minor variations for $k>1000$. 

\begin{table}[t]
 \small
 \centering
  \begin{tabular}{ll|l|l}
	Graph & Method &  RG & TAC09$_{200}$\\
    \hline 
\textsc{Hr} & NGD                                   & 81.8 $\ddagger$           & 57.5$\dagger$ \\ 
\textsc{Hr} & {\textsc{ppr}} (1 iter.)       & 43.4 $\dagger$ $\ddagger$ & 60.5$\dagger$ $\ddagger$\\ 
\textsc{Hr} & {\textsc{ppr}} (2 iter.)       & 78.3 $\ddagger$           & 66.0$\dagger$ $\ddagger$\\ \hline
\textsc{Hr} & {\textsc{ppr}} \emph{default}  & \textbf{88.4}             & \textbf{68.5} \\          
    \hline 
  \end{tabular}
  \caption{Result when using single links, compared to the use of the full graph on development data. 
           We reimplemented NGD. $\dagger$ for stat. signif. difference with \textsc{ppr}.
           $\ddagger$ for stat. signif. using all datasets. }
  \label{table:link-vs-graph}
\end{table}

\begin{table}[t]
 \small
 \centering
  \begin{tabular}{lll|l|l}
Graph & Method &	Year & RG         & TAC09$_{200}$\\
    \hline 
\textsc{Hr} & \textsc{ppr} \emph{default} &   2010     &       {86.3}   & 68.5 \\
\textsc{Hr} & \textsc{ppr} \emph{default} &    2011     & 85.6          & \textbf{70.5} \\
\textsc{Hr} & \textsc{ppr} \emph{default} &    2013     & \textbf{88.4} & 68.5 \\      
    \hline 
  \end{tabular}
  \caption{\textsc{ppr} using different Wikipedia versions}
  \label{table:wiki-versions}
\end{table}

\textbf{Is the full graph helping?} When the \textsc{ppr} algorithm
does a single iteration, we can interpret that it is ranking all
entities using direct links. When doing two iterations, we can loosely
say that it is using links at distance two, and so on. Table
\ref{table:link-vs-graph} shows that \textsc{ppr} is able to take
profit from the full graph well beyond 2 iterations, specially in
relatedness. These results were confirmed in the full set of datasets, with
statistically significant differences in all cases.

In addition, we reimplemented the relatedness and NED algorithms based
on NGD over direct links \cite{Milne08aneffective,Milne2008}, allowing
to compare them to \textsc{ppr} on the same experimental
conditions. We first developed the relatedness algorithm\footnote{In
  order to replicate the NGD relatedness algorithm, we checked the
  open source code available, exploring the use of inlinks and
  outlinks and the use of maximum pairwise article relatedness. We
  also realized that the use of priors (``commonness'' according to
  the terminology in the paper) was hurting, so we dropped it. We
  checked both reciprocal and unidirectional versions of the hyperlink
  graph, with better results for the reciprocal graph.}. Table
\ref{table:link-vs-graph} reports the best variant, which outperforms
the 0.64 on RG reported in their paper. We followed a similar
methodology for NED\footnote{We checked both reciprocal and undirected
  graphs with similar results, combined with prior (similar results),
  weighted terms in the context (with improvement) and checked the use
  of ambiguous mentions in the context (marginal
  improvement). Reported results correspond to reciprocal, combination
  with prior, weighting terms and using only monosemous mentions. }.
Table \ref{table:link-vs-graph} shows the results for NGD, which
performs worse than \textsc{ppr}. This trend was confirmed on the full
set of datasets for relatedness and NED with statistical significance
in all cases except KORE, which is the smallest NED dataset. Figure
\ref{fig:example-show-disamb} illustrates why the use of longer paths
is beneficial. In fact, NGD returns 0.14 for \texttt{B\&I\_Lions} and
0.13 for \texttt{Highveld\_Lions}, but \textsc{ppr} correctly returns
0.05 and 0.75, respectively.

\setlength{\tabcolsep}{3pt}
\begin{table*}[t]
 \small
 \centering
  \begin{tabular}{l|lc|rl|rl|rl|rl|rl}
\multicolumn{1}{c}{}                     & \multicolumn{2}{c}{Source} & \multicolumn{2}{c}{RG}  & \multicolumn{2}{c}{353} & \multicolumn{2}{c}{TSA} & \multicolumn{2}{c}{MC}& \multicolumn{2}{c}{KORE} \\ %
\hline 
\cite{ponzetto11a}                       & Wiki11               & c  &   & 75.0*          &   &                &   &               &   &               &   & \\           
\cite{DBLP:journals/ai/NastaseS13}       & Wiki13               & ci &   & 67.0           &   &                &   &               &   &               &   & \\           
\cite{milne_open-source_2013}            & Wiki13               & la &   & 69.5r          &   & 59.7r          &   & 35.8r         &   & 77.2r         &   & 65.9r \\           
\cite{yeh-EtAl:2009:TextGraphs4}         & Wiki09               & g  &   &                &   & 48.5           &   &               &   &               &   & \\           

\textbf{\textsc{ppr}} \emph{default} \textsc{Hr}     & \textbf{Wiki13}      & g  & 0 & \textbf{88.4}* & 1 & \textbf{72.8}  & 1 & \textbf{64.1} & 1 & \textbf{81.0} & 1 & \textbf{66.2} \\       
\hline 
\cite{AGIRRE10.534}                      & WNet                 & g  & 1 & \textbf{86.2r}          &   & 68.5           &   & 45.4r         & 3 & 85.2r         &   & \\           
\cite{DBLP:journals/jair/TsatsaronisVV10}& WNet                 & g  &   & 86.1           &   & 61.0           &   &               &   &               &   & \\           
\cite{DBLP:conf/aaai/NavigliP12}         & WNet+Wiki12 (cl)     & g+CL  &   &                &   & 65.0           &   &               & 1 & \textbf{90.0} &   & \\           
\cite{Pilehvaretal:2013}                 & WNet+Wiki13          & g  &   & 86.8*          &   &                &   &               &   &               &   & \\           
\textbf{\textsc{ppr}} \emph{default} \textsc{Hr}     & \textbf{Wiki13}      & g  & 0 & \textbf{88.4}* & 2 & \textbf{72.8}  & 1 & \textbf{64.1} & 4 & \textbf{81.0} & 1 & \textbf{66.2} \\       
\textbf{\textsc{ppr}} \emph{default} \textsc{Hr}     & \textbf{WNet+Wiki13}& g  & 0 & \textbf{91.8}* & 1 & \textbf{78.5}  & 2 & \textbf{62.9} & 2 & \textbf{87.6} & 1 & \textbf{66.2} \\           %
\hline 
\cite{gabrilovich2007csr}                & Wiki07               & t  &   & 82.0           &   & 75.0           &   & 59.0          &   & 73.0          &   & \\           
\cite{hoffart_kore:_2012}                & Wiki12               & t  &   &                &   &                &   &               &   &               & 0 & 69.8* \\           
\cite{Yazdani:2013:CTS:2405838.2405916}  & Freebase             & gt &   &                &   & 70.0*          &   &               &   &               &   & \\           
\cite{Radinsky:2011:WTC:1963405.1963455} & Time                 & C  &   &                & 1 & \textbf{80.0}  & 1 & \textbf{63.0} &   &               &   & \\           
\cite{baroni2014comparsion}              & Corpus               & C  &   & 84.0*          &   & 71.0           &   &               &   &               &   & \\                                           
\cite{agirre2009naacl}                   & WNet+Corpus       & Cg+SUP & 0 & \textbf{96.0}x &   & 78.0x          &   &               &   &               &   & \\           
\cite{milne_open-source_2013}            & Wiki13            & la+SUP &   & 83.5r          &   & 74.0x          &   & 52.8r         &   & 81.3r         & 1  & \textbf{66.5}r\\     
\textbf{\textsc{ppr}}  \emph{default} \textsc{Hr}    & \textbf{WNet+Wiki13} & g  & 0 & \textbf{91.8}* & 2 & \textbf{78.5}  & 2 & \textbf{62.9} & 2 & \textbf{87.6} & 2 & \textbf{66.2} \\           %
\hline 
  \end{tabular}
  \caption{Spearman results for relatedness systems. 
    The source column includes \textbf{codes} for information used (\textbf{t} for article text, \textbf{l} for direct hyperlinks, \textbf{g} for hyperlink graph, \textbf{c} for categories, \textbf{i} for infoboxes, \textbf{a} for anchor text) and other information sources (\textbf{CL} for crosslingual links, \textbf{C} for corpora, \textbf{SUP} for supervised Machine Learning). 
    The results include the following codes: \textbf{*} for best reported result among several variants, \textbf{x} for cross-validation result, \textbf{r} for third-party system ran by us. 
    We also include the rank of our \textsc{ppr} system in each group or rows, including the systems above it (excluding * and x systems, which get rank 0 if they are top rank). 
  }
  \label{table:simSOTA}
\end{table*}

\begin{table*}[t]
 \small
 \centering
  \begin{tabular}{l|lc|rl|rl|rl|rl|rl}
\multicolumn{1}{c}{System}                   & \multicolumn{2}{c}{Source}                   & \multicolumn{2}{c}{TAC2009} & \multicolumn{2}{c}{TAC2010}  & 
                                                      \multicolumn{2}{c}{TAC2013} & \multicolumn{2}{c}{AIDA}     & \multicolumn{2}{c}{KORE50} \\       
\hline                                                                                                             
MFS baseline                                  & Wiki13 & l      &   & 68.3            &   & 73.7          &   & 72.7          &   & 69.0          &   & 36.4  \\                            
\cite{guo_graph-based_2011}                   & Wiki10 & l      & 1 & \textbf{74.0}   &   & 74.1          &   &               &   &               &   &        \\                           
\cite{milne_open-source_2013}                 & Wiki13 & la     &   & 57.4r           &   & 58.5r         &   & 37.1r         &   & 56.0r         &   & 35.7r  \\                           
\cite{DBLP:journals/jair/GarciaAF14}          & Wiki12 & l      &   &                 &   & 76.6          &   &               &   &               &   &         \\
\textbf{\textsc{ppr}}  \emph{default} \textsc{Hr}         & Wiki13 & g      & 0 & \textbf{78.8}*  & 1 & \textbf{83.6} & 1 & \textbf{81.7} & 1 & \textbf{80.0} & 1 & \textbf{60.8}  \\                   
\hline                                          
\cite{Moro:2014:ELmeetsWSD}             & WNet+Wiki13  & g+CL   &   &                 &   &               &   &               & 1 & \textbf{82.1} & 1 & \textbf{71.5}  \\
\textbf{\textsc{ppr}}  \emph{default} \textsc{Hr}         & Wiki13 & g      & 0 & \textbf{78.8}*  & 1 & \textbf{83.6} & 1 & \textbf{81.7} & 2 & \textbf{80.0} & 2 & \textbf{60.8}  \\                   
\hline                                          
\cite{DBLP:conf/eacl/BunescuP06}              & Wiki11 & tc     & 0 & \textbf{83.8}ra* &   & 68.4ra        &   &               &   &               &   &        \\                           
\cite{Cucerzan2007}                           & Wiki11 & tc     & 0 & \textbf{83.5}ra*  &   & 78.4ra        &   &               &   & 51.0ro        &   &        \\                           
\cite{Hachey:2011:GNE:2050963.2050980}        & Wiki11 & tcg    &   &                 &   & 79.8*         &   &               &   &               &   &        \\                           
\cite{hoffart_kore:_2012}                     & Wiki12 & t      &   &                 &   &               &   &               & 0 & 81.8*         & 0 & 64.6* \\              
\cite{Hoffart11}                              & Wiki11 & tli+SUP &   &                 &   &               &   &               & 0 & 81.8*         &   &        \\                           
\cite{milne_open-source_2013}                 & Wiki13 & la+SUP  &   & 57.5r           &   & 63.4r         &   & 40.0r         &   & 55.6r         &   & 37.1r    \\                         
Best TAC KBP system                           & ---  & ---      & 1 & \textbf{76.5}   &   & 80.6          &   & 77.7          &   &               &   &        \\                           
\textbf{\textsc{ppr}}  \emph{default} \textsc{Hr}         & Wiki13 & g      & 0 & \textbf{78.8}*   & 1 & \textbf{83.6} & 1 & \textbf{81.7} & 2 & \textbf{80.0} & 2 & \textbf{60.8}  \\                   
\hline
  \end{tabular}
  \caption{Accuracy of NED systems, using the same codes as in in Table \ref{table:simSOTA}. 
    Some early systems have been re-implemented and tested by others: 
    \textbf{ra} for \cite{Hachey2012}, \textbf{ro} \cite{Hoffart11}. 
    We report rank of our \textsc{ppr} system in each group or rows, including systems above (excluding * systems, which get rank 0 if they are top rank). 
  }
  \label{table:ned_testresults}
\end{table*}

\textbf{How important is the Wikipedia version?} Table
\ref{table:wiki-versions} shows that the versions we tested are not
affecting the results dramatically, and that using the last version does not yield better results in NED.
Perhaps the larger size and 
 number of hyperlinks of newer versions would only affect new articles and rare
articles, but not the ones present in TAC09$_{200}$. We kept using 2013 for test.

\textbf{What is the efficiency of the algorithm?}  The initialization takes
around 5 minutes\footnote{Time measured in a single server with Xeon E7-4830
  8 core processors, 2130 MHz, 64 GB RAM.}, where most of the time is spent
loading the dictionary into memory, 4m50s. Using a database instead,
initialization takes 10s. Memory requirements for \textsc{Hr} were 4.7 Gb,
down to 1.1 Gb when using the database. The main bottleneck of our system is
the computation of Personalized PageRank, each iteration taking around 0.60
seconds. We are currently checking fast approximations for Pagerank, and
plan to improve efficiency.

\section{Comparison to related work}
\label{sec:comp-relat-work}

In the previous section we presented several results on the same
experimental conditions. We now use the graph and parametrization
which yield the best results on development (default parameters with \textsc{Hr}). Comparison to the state of the art is complicated by many
systems reporting results on different datasets, which causes the
tables in this section to be rather sparse. The comparison for
relatedness is straightforward, but, in NED, it is not possible to
factor out the impact of the candidate generation step. Given the fact
that our candidate generation procedure is not particularly
sophisticated, we don't think this is a decisive factor in favour of
our results.

Table \ref{table:simSOTA} and \ref{table:ned_testresults} report the
results of the best  systems on both tasks. Given that several
systems were developed on test data, we also report our results on RG
and TAC2009, marking all such results (see caption of tables for
details). We split the results in both tables in three sets: top rows
for systems using link and graph information alone, middle rows for
link- and graph-based systems using WordNet and/or Wikipedia, and bottom
rows for more complex systems. We report the results of our system
repeatedly in each set of rows, for easier comparison. Our main focus
is on the top rows, which show the superiority of our results
with respect to other systems using Wikipedia links and graphs. The
middle and bottom rows show the relation to the state of the art.

For easier exposition, we will examine the results by row section
simultaneously on relatedness and NED. The \textbf{top rows} in Table
\ref{table:simSOTA} report four relatedness systems which have already
been presented in Sect.  \ref{sec:previous-work}, showing that our
system is best in all five datasets. Note that the
\cite{milne_open-source_2013} row was obtained running their publicly
available system with the supervised Machine Learning component turned off (see
below for the results using SUP). The top rows of table
\ref{table:ned_testresults} report the most frequent baseline (as
produced by our dictionary) and three link-based systems
(cf. Sect. \ref{sec:previous-work}), showing that our method is best
in all five datasets.  These results show that the use of the full
graph as devised in this paper is a winning strategy.

The relatedness results in the \textbf{middle rows} of Table \ref{table:simSOTA} include several systems using WordNet and/or Wikipedia (cf. Sect. \ref{sec:previous-work}), including the system in \cite{AGIRRE10.534}, which we
 run out-of-the-box with default values.  To
date, link-based systems using WordNet had reported stronger results than their
counterparts on Wikipedia, but the table shows that our Wikipedia-based results are the strongest on
all relatedness datasets but one (MC, the smallest dataset, with only 30 pairs). In addition, the table shows our
results when combining random walks on Wikipedia and
WordNet\footnote{We multiply the scores of \textsc{Ppr} on Wikipedia and WordNet.}, which yields improvements in most datasets. In the counterpart for NED in Table \ref{table:ned_testresults}, Moro et al. \shortcite{Moro:2014:ELmeetsWSD} outperform our system, specially in the smaller KORE (143 instances), but note that they use a richer graph which combines WordNet, the English Wikipedia and hyperlinks from other language Wikipedias.

Finally, the \textbf{bottom rows} in both tables report the best
systems to date. For lack of space, we cannot review systems not using
Wikipedia links. Regarding relatedness, we can see that our
combination of WordNet and Wikipedia would rank second in all
datasets, with only one single system (based on corpora) beating our
system in more than one dataset
\cite{Radinsky:2011:WTC:1963405.1963455}. Regarding NED, our system
ranks first in the TAC datasets, including the best systems that
participated in the TAC competitions
\cite{varma2009,Lehmann2010,cucerzan2013tac}, and second to
\cite{Moro:2014:ELmeetsWSD} on AIDA and KORE.

\section{Conclusions and Future Work}
\label{sec:concl-future-work}

This work departs from previous work based on Wikipedia and derived
resources, as it focuses on a single knowledge source (links in
Wikipedia) with a clear research objective: given a well-established
random walk algorithm we explored which sources of links and filtering
methods are useful, contrasting the use of the full graph with respect
to using just direct links. We follow a clear
development/test/analysis methodology, evaluating on a extensive range
of both relatedness and NED datasets. All software and data are
publicly available, with instructions to obtain out-of-the-box
replicability\footnote{\url{http://ixa2.si.ehu.es/ukb/README.wiki.txt}}.

We show for the first time that random walks over the full graph of
links improve over direct links. We studied several variations of
sources of links, showing that non-reciprocal links hurt and that
the contribution of the category structure and relations in infoboxes is
residual. This paper sets a new state-of-the-art for systems based on
Wikipedia links on both word relatedness and named-entity
disambiguation datasets. The results are close to those of the best
combined systems, which specialize on either relatedness or
disambiguation, use several information sources and/or
supervised machine learning techniques. This work shows that a careful
analysis of varieties of graphs using a well-known random walk
algorithm pays off more than most ad-hoc algorithms proposed up to
date.

For the future, we would like to explore ways to filter
out informative hyperlinks, perhaps weighting edges according to their
relevance, and would also like to speed up the random-walk computations. 

This article showed the potential of the graph of hyperlinks. We would
like to explore combinations with other sources of information and
algorithms, perhaps using supervised machine learning. For
relatedness, we already showed improvement when combining with random
walks over WordNet, but would like to explore tighter integration
\cite{Pilehvaretal:2013}. For NED, local methods
\cite{RatinovRDA11,Han2011}, global optimization strategies based on
keyphrases in context like KORE \cite{hoffart_kore:_2012} and doing
NED jointly with word sense disambiguation \cite{Moro:2014:ELmeetsWSD},
all are complementary to our method and thus promising directions.

\section*{Acknowledgements}

This work was partially funded by MINECO (CHIST-ERA READERS project --
PCIN-2013-002- C02-01) and the European Commission (QTLEAP --
FP7-ICT-2013.4.1-610516).  Ander Barrena is supported by a PhD grant
from the University of the Basque Country.

\end{document}